%% file: main.tex
\documentclass[runningheads]{llncs}

\usepackage[year=2024,ID=5623]{eccv}
\usepackage[pagebackref,breaklinks,colorlinks,citecolor=eccvblue]{hyperref}
\usepackage{eccvabbrv}
\usepackage[accsupp]{axessibility}

\usepackage{multirow}
\usepackage{indentfirst}
\usepackage{xcolor}

\usepackage{colortbl}
\usepackage{tabularx}
\usepackage{tabulary,multirow,overpic,xcolor,subfloat}
\usepackage{orcidlink}
\usepackage{booktabs}
\usepackage{graphicx}

\begin{document}
\title{EventRR: Event Referential Reasoning \\ for Referring Video Object Segmentation}

\def\methodName{Event Referring Reasoning}
\author{Huihui Xu\inst{1,2,3} \and
Jiashi Lin\inst{1,4} \and
Haoyu Chen \inst{2}\and
Junjun He \inst{1}\and
Lei Zhu\inst{3,2}\thanks{Corresponding author} }

\authorrunning{H.~Xu et al.}

\institute{
Shanghai Artificial Intelligence Laboratory \and
The Hong Kong University of Science and Technology (Guangzhou)\\
\email{leizhu@ust.hk} \and
The Hong Kong University of Science and Technology\and
Northwestern Polytechnical University
}




\maketitle

\begin{abstract}
Referring Video Object Segmentation (RVOS) aims to segment out the object in a video referred by an expression. 
Current RVOS methods view referring expressions as unstructured sequences, neglecting their crucial \textit{semantic structure} essential for referent reasoning.
Besides, in contrast to image-referring expressions whose semantics focus only on object attributes and object-object relations, video-referring expressions also encompass \textit{event attributes} and \textit{event-event temporal relations}. This complexity challenges traditional structured reasoning image approaches.
In this paper, we propose the \textbf{Event Referential Reasoning (EventRR)} framework. EventRR decouples RVOS into object summarization part and referent reasoning part. The summarization phase begins by summarizing each frame into a set of bottleneck tokens, which are then efficiently aggregated in the video-level summarization step to exchange the global cross-modal temporal context.
For reasoning part, EventRR extracts semantic eventful structure of a video-referring expression into highly expressive \textbf{Referential Event Graph (REG)}, which is a single-rooted directed acyclic graph. 
Guided by topological traversal of REG, we propose \textbf{Temporal Concept-Role Reasoning (TCRR)} to accumulate the referring score of each temporal query from REG leaf nodes to root node. Each reasoning step can be interpreted as a question-answer pair derived from the concept-role relations in REG. Extensive experiments across four widely recognized benchmark datasets, 
show that EventRR quantitatively and qualitatively outperforms state-of-the-art RVOS methods. Code is available at \url{https://github.com/bio-mlhui/EventRR}.
\end{abstract}
\section{Introduction}
\label{sec:intro}
\begin{figure}[htp]
\centering
\includegraphics[width=1.0\linewidth]{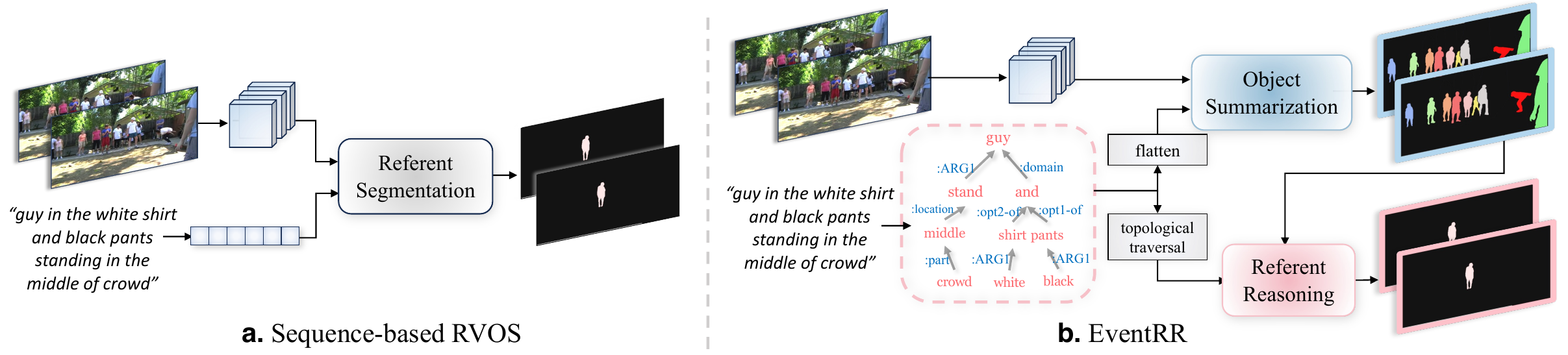}
\caption{Comparing the pipelines between state-of-the-art sequenced-based RVOS methods and EventRR.}
\label{fig:framework}
\end{figure} 
A referring expression naturally exposes a compositional logical procedure to iteratively filter out background objects and locate the referent object. Specifically, semantics of an image-referring expression include \textbf{(1)} object attributes ($\text{attribute}\stackrel{\text{ }}{\longrightarrow}\text{object}$), such as appearance ("the blue shirt") and motion ("the walking man"); \textbf{(2)} object-object relations ($\text{object}\stackrel{\text{relation}}{\longrightarrow}\text{object}$), such as spatial position (“the man to the left of the chair") and predicates ("man drinking water"). From this perspective, an effective  referring model should not only learn fine-grained features to distinguish different objects in the same image/video, but also utilize the semantic structure within the expression to reason the referent object~\cite{xu2025medground,wang2024language}.
Most importantly, we claim that apart from image-referring semantics of \textbf{(1)} object attributes and \textbf{(2)} object-object relations, a typical video-referring expression in existing RVOS datasets also includes \textbf{(3)} event attributes, such as motion source and destination ("moving from left to right"); \textbf{(4)} event-event temporal relations, such as ("walking then running").

Many parsing-tree based Referring Image Object Segmentation (RIOS) methods \cite{Linguistic,huang2020referring, Bottom-up} have been proposed to devise the semantic structure in the image-referring expression. Recently, BUSNet\cite{Bottom-up} proposes to extract the image-referring expression into an Image Scene Graph (ISG) \cite{johnson2015image, krishna2017visual} and iteratively refines the backbone feature maps guided by the ISG structure. SVG-Tree \cite{svgtree} proposes to parse the sentence into a binary Recursive Grounding Tree (RvG-Tree) and follows the divide-and-conquer strategy with score/feature node classification to accumulate the referring score of each ROI feature. Generally, existing referring image methods can be classified to Dependency-based \cite{svgtree,Linguistic}, Attention-based \cite{cmpc, huang2020referring}, and Image Scene Graph based (ISG-based)\cite{yang2020graph, Bottom-up, wu2023grounded}. 

Dependency-based \cite{svgtree,Linguistic} methods devise the \textit{syntactic} constituency dependency \cite{manning2014stanford} structure, but ignore the more informative \textit{semantic} information in image/video-referring expressions. Attention-based methods \cite{cmpc, huang2020referring} dynamically build nodes graph guided by the multi-modal feature attention map, which is hidden and has no interpretable semantic meanings. ISG-based \cite{yang2020graph, Bottom-up, wu2023grounded} methods parse the expression into an Image Scene Graph utilizing a text-to-ISG parser \cite{schuster2015generating}. Although ISG captures the high-level semantics in image-referring expressions, ISG nodes can only represent objects, object attributes, and binary object-object relations, and ISG edges have no semantic meanings (A classical ISG example can be found in Fig.4\cite{krishna2017visual}). They are not expressive enough to represent \textbf{(3)} event attributes and \textbf{(4)} event-event relations, such as "moving from left to right", "walk then run", which frequently appear in existing RVOS datasets. Although Video Scene Graph Generation (VSGG, video to VSG) \cite{vsg1, vsg2} is proposed, VSG nodes and edges cannot directly represent semantics like destination, source, duration for an event. Above all, parsing a video-referring expression (one item) to VSG (sequence of graph items by frame) is also ill-conditioned.

Moreover, existing parsing-tree RIOS methods \cite{Linguistic, huang2020referring, Bottom-up} or parsing-tree grounding methods \cite{ wu2023grounded, yang2020graph} either iteratively refine the backbone feature map or require a set of off-the-shelf object proposals as input. They cannot be applied to SOTA DETR\cite{detr}-based segmentation frameworks like Mask2Former\cite{cheng2022masked}. The combination between parsing-tree based referent reasoning and object queries with end-to-end bipartite matching learning \cite{detr} has not been explored. 

Above limitations motivate us to propose the \textbf{Event Referential Reasoning (EventRR)} framework.  As far as we know, EventRR is the first parsing-tree based RVOS method. As shown in Figure \ref{fig:framework}(a), sequence-based RVOS methods \cite{referformer, soc, spectrum, r2vos, mttr, mevis, tang2023temporal, r2vos, wu2023onlinerefer} ignore instance-level discrimination and treat RVOS as vanilla foreground/background segmentation. Specifically, SOC\cite{soc}, SgMg\cite{spectrum}, TCD\cite{tang2023temporal} and Referformer\cite{referformer} initialize decoder object queries with sentence-level language features~\cite{roberta} and a simple binary classification head is devised to directly output foreground probability of each query. CMPC-V\cite{cmpc} uses soft attention on sequence features to extract the entity, attribute, and relation information in the expression into single vectors. In all, above methods treat the expression as a structureless sequence and ignore its compositional semantic structure which is crutial for referent reasoning. Although these methods achieve satisfactory performance, their model lacks interpretability and may fail when the expression contains more complicated hierarchical semantic structure, which, as shown in \cite{lost}, cannot be handled well by current large-scale pre-trained language models \cite{roberta}.  

As shown in Figure\ref{fig:framework}(b), EventRR extracts the semantic structure in a video-referring expression into the \textbf{Referential Event Graph (REG)}, which is a single-rooted DAG and induces clear reasoning procedure. EventRR decouples RVOS into the object summarization part and the referent reasoning part. The object summarization part uses a set of temporal queries to summarize the instance-level information in the video and is obligated to highlight all relevant objects in the expression. Different from other methods \cite{soc, referformer, mttr} where the contextualized expression features \cite{roberta} are used, EventRR fuses visual features with the highly abstracted semantic concept, role features \cite{bai2022graph}, which can help EventRR focus on the most informative semantic information. For Referent Reasoning, we propose \textbf{Temporal Concept-Role Reasoning (TCRR)} which utilizes the topological traversal in REG to recursively update the referring score of each temporal query in a bottom-up manner. During inference, EventRR takes the mask prediction of the query with the highest referring score as output.

In summary, the contributions of this paper are:
\begin{itemize}
\item To capture the semantics of event-event relations and event attributes in video-referring expressions, we propose a new structured representation called Referential Event Graph (REG) to enable RVOS models to understand the complicated semantic structure in a video referring expression. 
\item We design the Temporal Concept-Role Reasoning (TCRR) module to explicitly utilize the compositional semantic structure in the expression for referent reasoning, which accumulates the referring score of each temporal query in a bottom-up manner guided by the topological traversal of REG. As far as we know, we are the first to explore integrating  parsing-tree based referent reasoning to DETR-based bipartite matching learning.
\item Extensive experiments demonstrate that EventRR achieves state-of-the-art performance on Ref-Youtube-VOS, DAVIS17-RVOS, A2D-Sentences and JHMDB-Sentences.
\end{itemize} 
\section{Related Work}
\subsection{Referring Video Object Segmentation}
Referring Video Object Segmentation (RVOS) aims to segment out the object in a video referred by an expression. MTTR \cite{mttr} introduces DETR\cite{detr}-based framework into RVOS, which uses self-attention based multi-modal transformer to fuse concatenated visual and textual features. A set of learnable queries are used to iteratively refine mask predictions at each decoder layer. ReferFormer \cite{referformer} proposes to condition queries with sentence-level textual features. SOC \cite{soc} proposes the Visual-Linguistic Contrastive Loss to contrast the pooled sentence-level features with the matched referent query. OnlineRefer\cite{wu2023onlinerefer} proposes inter-frame query propagation and extends RVOS to online setting. HTML\cite{html} proposes hybird temporal training to efficiently learn the cross-modal temporal context correspondence. R2VOS\cite{r2vos} proposes to improve segmentation robustness by optimizing the dual expression reconstruction task, which mitigates the false-alarm problem. SgMg\cite{spectrum} proposes Spectrum-guided fusion to efficiently facilitate global interaction. In all, existing state-of-the-art RVOS methods treats the referring expression as unstructured sequence and ignore the semantic structure. As far as we know, no parsing-tree based methods are proposed for RVOS. Moreover, RVOS has been extended to medical domain~\cite{wang2024video,ning2025retinalogos,xu2024lgrnet,xu2025medground,wang2025serp,tian2023deep,wu2024rainmamba}.
\subsection{Abstract Meaning Representation}
Proposed and augmented in \cite{amr, spatial_amr}, Abstract Meaning Representation (AMR) captures "who is doing what to whom" in any sentence. Recently, text to AMR parsing \cite{text_to_amr_1, text_to_amr_2, text_to_amr_3} has achieved great success, which makes it possible to use AMR to conduct symbolic reasoning \cite{amr_reason1, amr_reason2, amr_reason3} and question answering\cite{amr_qa1}. Next, we give a brief preliminary of AMR for clarification.

For an English sentence, its AMR is a rooted directed acyclic graph. Node in an AMR is called \textit{\textbf{concept}}, whose vocabulary contains PropBank frames \cite{kingsbury2002treebank} or free-form English words. Each PropBank frame represents a sense of predicate and assigns meaning for its different arguments. One example is "walk-01", where its "ARG0" means "walker", its "ARG1" means "entity being walked", and its "ARG2" means "location walked". Different predicates could have the same lemma. For example, "run-02" represents running motion, while "run-01" can represents "running a company". Edge in AMR is called \textit{\textbf{role}}. It connects two concepts and assigns one semantic meaning to their relations. The edge vocabulary includes PropBank arguments "ARGx" (core roles) and highly expressive roles (non-core roles), such as ":time", ":destination", ":duration", and ":source". For example, AMR can represent "walk then run" as "$\text{walk}\stackrel{\smash{\text{:opt1}}}{\leftarrow}$and$\stackrel{\smash{\text{:opt2}}}{\rightarrow}$run$\stackrel{\smash{\text{:time}}}{\rightarrow}\text{then}$. Moreover, SpatialAMR \cite{spatial_amr} augments node vocabulary with concepts like "left-20" to treat spatial position relation as a predicate. More detailed examples, rules, and role definitions of AMR can be found in \cite{amr_website}.

We take inspiration from the highly expressive power of AMR and help RVOS models to comprehensively understand the semantics of event attributes and event-event relations in video-referring expressions.
\section{Method}
\begin{figure*}[t]
\centering
\includegraphics[width=\textwidth]{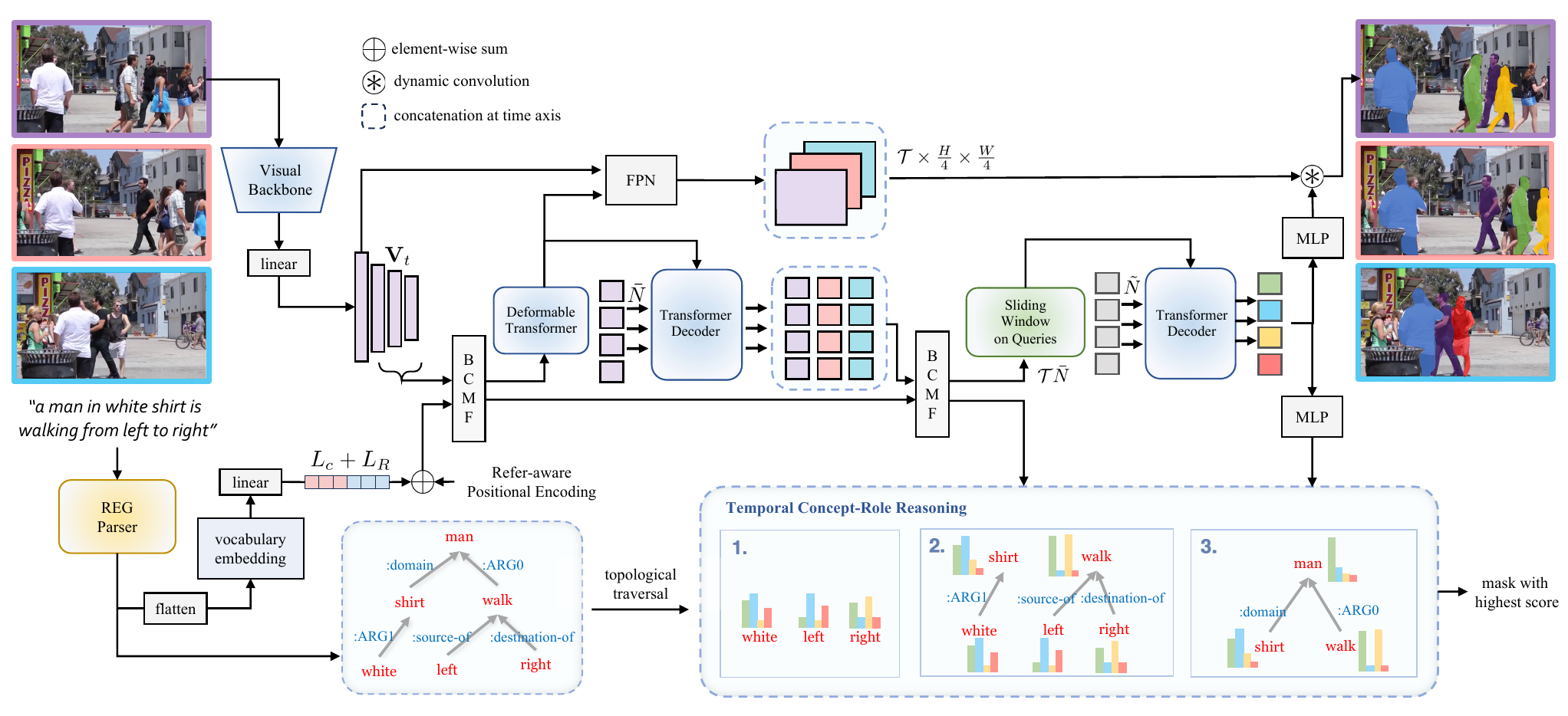}
  \caption{Event Referential Reasoning (EventRR) Framework. Our model takes a video clip and a video-referring expression as input. The visual side is composed of multi-scale feature extraction, frame-level object summmarization, and video-level object summarization. On the linguistic side, the Referential Event Graph (REG) parser first extracts the compositional semantic structure of the given expression into a REG graph, which induces a clear referent reasoning procedure. A set of frame queries first summarize each frame, and all frame queries are path-connected to each other by the Sliding Window on Queries (SWQ) module to exchange the global temporal context. We use Bilateral Cross-Modal Fusion (BCMF) to early fuse the REG concept-role features with the multi-scale visual features and all frame queries to highlight all instances mentioned in the expression. Guided by the topological traversal of REG, the Temporal Concept-Role Reasoning (TCRR) takes the temporal queries as input and recursively update the referring score of each query at each reasoning step. During inference, prediction of the temporal query with the highest referring score is output.}
\label{fig:arch}
\end{figure*}
Given a video $V=\{V^t\}_{t=1}^\mathcal{T}$ containing a set of objects $O=\{O_i\}_{i=1}^{N_o}$ and a referring expression $G$ of free-form English words, RVOS assumes $G$ could unambiguously distinguish the referent $O_r$ from other objects, and expects the model to output its segmentation mask $\mathbf{M}_r\in \{0,1\}^{\mathcal{T}\times H\times W}$. 

\subsection{Visual Feature Extraction}\label{sec:visual_extraction}
For each frame $V^t$, we use a visual backbone such as Swin Transformer \cite{swin} to extract a set of multiscale features $\mathbf{V}^t = \{\mathbf{V}^t_i\in \mathcal{R}^{ H_iW_i\times c_i}\}$, where $i\in\{1,2,3,4\}$, and $H_i$, $W_i$, $c_i$ denotes the spatial height, width and dimension of the feature map, respectively. For each scale, a linear layer is applied to transform from $c_i$ into $d$ dimensional space.

\subsection{Referential Event Graph}\label{sec:semantic_extraction}
We first use a transition-based AMR parser \cite{amr_parser} to get the AMR of the expression. Since the AMR annotation process \cite{amr} does not specify the focus of the sentence, the AMR root node does not often correspond to the referent main concept. Moreover, since co-reference may appear, rings will appear in the adjacency-based graph representation, which fails topological traversal. To solve the first issue, we aim to use the syntactic structure \cite{manning2014stanford} of the sentence to choose the referent concept. To solve the second issue, we use depth-first search to incrementally build the directed acyclic graph starting from the root. We call the final graph as Referential Event Graph (REG), which induces a clear referent reasoning process for RVOS models. \footnote{Due to space limitation, we illustrate detailed AMR to REG transformations and several examples in the Supplementary.}

To get features of the semantic concepts and semantic roles, we use the vocabulary embedding of pre-trained AMRBART-Large \cite{bai2022graph} on AMR 3.0\cite{amr3}. AMRBART uses a shared vocabulary for concepts, roles, and free-form English words. For tokens outside the vocabulary, its feature is taken as the average of its BPE subtoken features. A linear layer is then used to transform into $d$ dimensional space. The final REG graph features are denoted as 
\begin{equation}
    \mathbf{G}=\{\mathbf{C}\in \mathcal{R}^{L_c\times d}, \mathbf{R}\in \mathcal{R}^{L_R\times d}, \mathbf{E}\in \mathcal{Z}^{L_R\times 2} \},
\end{equation} where $\mathbf{C}$ is concept features, $\mathbf{R}$ is role features, and each row in $\mathbf{E}$ denotes the source concept index and the target concept index of an edge.

\noindent\textbf{Refer-Aware Positional Encoding.} Different from other general graph types where nodes have no canonical positional information, each node in REG is an ascendant of the root concept. Intuitively, if the distance from a concept to the root is larger, the less information it may contribute to distinguishing the referent from the context. To encode this depth-aware positional information, we propose to add Refer-Aware Positional Encoding (ReferPE) to the concept features. Specifically, we set the max distance to 50. For each node, the learnable embedding corresponding to its shortest path distance to the root is added to its feature:
\begin{equation}
    \mathbf{C} = \mathbf{C} + \textbf{PE}^{refer}.
\end{equation}

\subsection{Object Summarization}
The object summarization part is composed of three components: frame-level summarization, video-level summarization, and Bilateral Cross-Modal Fusion (BCMF). 

\noindent\textbf{Frame-level Summarization.}\label{sec:perframe_obj}
The frame-level encoder-decoder adopts the same architecture in Mask2Former \cite{cheng2022masked}. Specifically, the frame encoder is a stack of 6 layers of Deformable Attention\cite{deform_detr} to fuse the last three multiscale features. A FPN \cite{lin2017feature} is used to pass the encoded high-level semantic information to the first scale feature. Finally, each frame is summarized into a set of frame queries $\bar{\mathbf{O}}^t\in\mathcal{R}^{\bar{N}\times d}$,
where $\bar{N}$ is the number of frame queries used by the frame decoder.

\noindent\textbf{Video-level Summarization.}\label{sec:swq} 
 Since the descriptive information in an expression may span multiple frames, models trained using a sampled small clip of the video may fail to learn the correspondence between the long context temporal visual information and referring information. To mitigate this issue, we propose to aggregate the long context temporal information based on the summarized frame queries. Instead of using vanilla self-attention which has quadratic complexity of the sequence length and may attend to redundant information among different frames, we use the Sliding Window Attention\cite{swin} as the base layer. In each encoder layer, frame queries $\{\bar{\mathbf{O}}^t\}_{t=1}^{\mathcal{T}}\in\mathcal{R}^{\mathcal{T}\times\bar{N}}$ are split into a set of windows of size $\frac{\mathcal{T}}{W}\times \bar{N}$ along the time axis. The self-attention is applied at each window in parallel. By shifting the window partition by $\frac{W}{2}$ between layers, the final encoded features can create interaction among far frames, which can efficiently aggregate long context temporal information.
 
 The temporal decoder uses a set of temporal queries to aggregate information of the last SWQ layer output:
 \begin{equation}
     \mathbf{\tilde{O}} = \textbf{Decode}(\mathbf{\{\bar{O}\}}^t) \in \mathcal{R}^{\tilde{N}\times d},
 \end{equation}
 where $\tilde{N}$ is the number of temporal queries used by the temporal decoder.

 \begin{figure}[ht]
  \centering
  \begin{minipage}[b]{0.45\textwidth}
    \includegraphics[width=\textwidth]{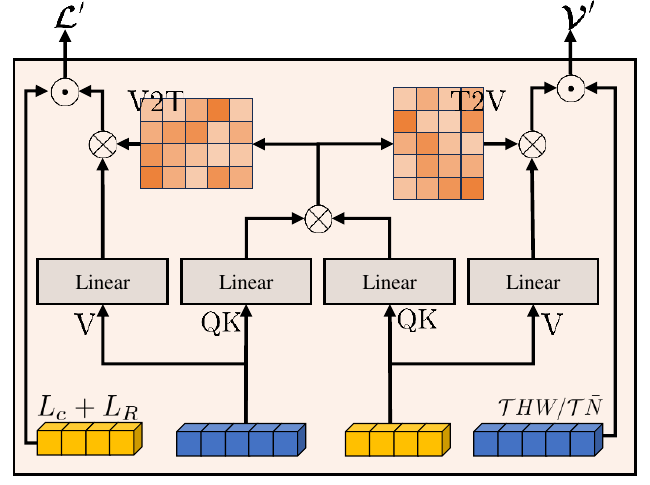}
    \caption{Module design of Bilateral Cross Modal Fusion (BCMF).}
    \label{fig:fusion_sup}
  \end{minipage}
  \hspace{-0.1mm}
  \begin{minipage}[b]{0.4\textwidth}
    \includegraphics[width=\textwidth]{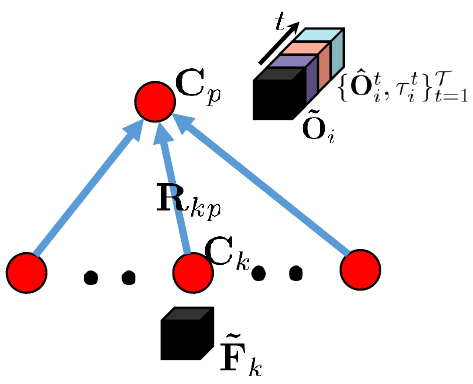}
    \caption{A generalized reasoning step in TCRR.}
    \label{fig:step}
  \end{minipage}
\end{figure}
\noindent\textbf{Bilateral Cross-Modal Fusion.}
As shown in Fig.\ref{fig:fusion_sup}, we design the Bilateral Cross-Modal Fusion (BCMF) module to interact information between two modalities. The first BCMF module takes multi-scale features as input, which is obligated to exchange spatial information between two modalities for each frame. The second BCMF module takes the all frame queries as input, which is obligated to exchange the cross-modal temporal information. In BCMF, the common un-normalized attention weight are first computed. Then, for T2V (V2T) where visual (text) part is the query feature, the weights are softmax normalized along text(vision) dimension and used to compute weighted sum of text(vision) features. Information flow in BCMF is bidirectional, in the sense that both the visual features and flattened REG features get updated.

\subsection{Temporal Concept-Role Reasoning.}


The temporal decoder summarizes the high-level semantic information in the video into a sequence of temporal queries. Moreover, for each temporal query, we choose $\mathcal{T}$ frame queries represent its frame-level summarization. Formally, we denote $\mathbf{\Gamma}_i\in \mathcal{R}^{\mathcal{T}\times \bar{N}}$ as the cross attention weights between temporal query $\tilde{O}_i$ and frame queries $\{\bar{O}^t\}_{t=1}^{\mathcal{T}}$ of the last SWQ layer. Then queries input to TCRR can be represented as 
\begin{equation}
  \mathbf{O}_i=[\mathbf{\tilde{O}}_i, \{\mathbf{\hat{O}}^t_i, \tau_i^t\}_{t=1}^\mathcal{T}], \quad i=1,...,\tilde{N},
\end{equation}
where by denoting $j=\mathop{\arg\max}_{j=1,...,\bar{N}}\mathbf{\Gamma}_i^{t,j}$, we have $\tau_i^t=\mathbf{\Gamma}_i^{t,j}$ and $\mathbf{\hat{O}}^t_i=\mathbf{\bar{O}}^t_j$.

\noindent\textbf{Reasoning Order.} We use Kahn's algorithm \cite{kahn1962topological}\footnote{We use the implementation in \href{https://docs.dgl.ai/en/0.2.x/generated/dgl.traversal.topological_nodes_generator.html}{DGL}} to get the topological traversal of the nodes in REG. The reasoning order derived from the topological traversal can guarantee all child nodes can pass their messages to the parent \textit{at the same reasoning step}. Moreover, since our graph is single-rooted DAG, when reasoning proceeds to some parent node, messages from its childs only come from the sub-graph rooted by it, which is also a single-rooted DAG. Therefore, we can interpret the final referring score of each node as the referring distribution given by the sub-expression corresponding to the sub-graph rooted by that node. We implement the bottom-up reasoning order as in \url{https://github.com/pyg-team/pytorch_geometric/issues/121#issuecomment-471630847}.

\noindent\textbf{Reasoning Step.} Each reasoning step can be generalized as in Figure \ref{fig:step}. The parent concept feature is denoted as $\mathbf{C}_p$, and we assume it has $K$ role connections with $K$ childs 
$\{C_k, R_{kp}\}_{k=1}^{K}$. For each node $C$, its unormalized referring score for object $O_i$ is denoted as ${\boldsymbol{\sigma}(C, O_i)}\in \mathcal{R}$.

Two components are used to compute each score: the Object-Concept Align (OCA), and the Temporal Referent-Context Align (TRCA).

\noindent\textbf{Object-Concept Align.} This component directly computes the alignment between the temporal query feature and the parent concept feature:
\begin{equation}
    \boldsymbol{\sigma_r}(C_p, O_i) = \boldsymbol{\omega_r}^T(\mathbf{W_r}^T\mathbf{\tilde{O}}_i\odot \mathbf{C}_p),
\end{equation}
where $\mathbf{W_r}\in \mathcal{R}^{d\times d}$ and $\boldsymbol{\omega_r}\in \mathcal{R}^d$.

\noindent\textbf{Temporal Referent-Context Align.} This component computes the alignment between the referent-context relation information of the video and that in the REG. Our design principle is to treat the semantic role of the REG edge as the relationship between the referent (parent concept) and the context (its sub-graph). It is composed as a sum of $K$ parts for each child, where each part is formulated as:
\begin{equation}
    \boldsymbol{\sigma_e}(C_p, C_k, O_i) = \sum_{t=1}^{\mathcal{T}} \tau_i^t \cdot \boldsymbol{\omega_e}^T(\mathbf{W_e}^T[\tilde{\mathbf{F}}_k, \mathbf{\hat{O}}_i^t]\odot \mathbf{R}_{kp})
\end{equation}
\begin{equation}
    \mathbf{\tilde{F}}_k = \text{softmax}_i(\boldsymbol{\sigma}(C_k, O_i))^T \mathbf{\tilde{O}}, i=1,...,\tilde{N}
\end{equation}
where $\mathbf{W}_e\in \mathcal{R}^{2d\times d}$, $\boldsymbol{\omega}_e\in \mathcal{R}^d$, $[\text{ }]$ is concatenation of two vectors, and $\text{softmax}_i$ is taking the softmax of $\tilde{N}$ scores.
The final grounding score for the parent node is 
\begin{equation}
    \boldsymbol{\sigma}(C_p, O_i) = \boldsymbol{\sigma_r}(C_p, O_i) + \sum_{k=1}^K \boldsymbol{\sigma_e}(C_p, C_k, O_i).
\end{equation}

\noindent\textbf{Interpretation as Question-Answer.} Each reasoning step can be interpreted as one Question-Answer pair. Taking left$\hspace{-0.3cm}\stackrel{\smash{\text{:destination-of}}}{\longrightarrow}\hspace{-0.3cm}$jump as an example, the OCA (Question, Answer) pair could be (What are you doing, "jump"). Each temporal query is asked ($\mathbf{R}=\mathbf{W_r}^T\mathbf{\tilde{O}}_i$), then the response-answer similarity is measured($\mathbf{R}\odot \mathbf{C}_p$). The TRCA (Question, Answer) pair could be (what "left" means to you, "destination"). The "left" information, i.e. $\tilde{\mathbf{F}}_k$, comes from the temporary reasoning result of its sub-graph. Since edge may represent event attributes (:destination, :source) or temporal event relations (:time), each temporal query is allowed to utilize its per-frame context $\{\mathbf{\hat{O}}_i^t\}_{t=1}^{T}$ to answer the question $\mathbf{R}^t=\mathbf{W_e}^T[\tilde{\mathbf{F}}_k, \mathbf{\hat{O}}_i^t]$, then per-frame response-answer similarity is weighted by per-frame attentive weights $\sum_{t=1}^{\mathcal{T}} \tau_i^t \cdot \boldsymbol{\omega_e}^T(\mathbf{R}^t\odot \mathbf{R}_{kp})$. 
\subsection{Training TCRR with Bipartite Matching}
For box and mask supervision, we use the multi-task loss:
\begin{equation}
    \lambda_{\text{mask}} L_{\text{mask}} + \lambda_{\text{dice}} L_{\text{dice}} + \lambda_{\text{giou}} L_{\text{giou}} + \lambda_{\text{L1}} L_{\text{L1}},
\end{equation}
where $L_{\text{mask}}$, $L_{\text{dice}}$, $L_{\text{giou}}$, and $L_{\text{L1}}$ denote the mask binary cross entropy loss, DICE loss \cite{dice}, GIoU loss \cite{giou} and box L1 loss, respectively. We use bipartite matching \cite{detr} to first find the query that best matches the referent, and then optimize its mask and box prediction to the ground truth.

\noindent\textbf{Pseudo-Referent Reasoning Loss (PseudoRRL).} We want the model to maximize the referring score of the temporal query whose mask prediction is the referent. Although we can not know which query attends to the referent, we can utilize the bipartite matching result and treat the query matched with the referent ($i$-th query) as $\textit{pseudo-referent}$ and maxmize its referring score. Specifically, the PseudoRRL loss $L_{\text{reason}}$ is the cross entropy between referring score of all queries (a vector of shape $\tilde{N}$) with a one-hot vector whose only the $i$-th position is 1. 

The final loss is: 
\begin{equation}
    \lambda_{\text{reason}} L_{\text{reason}} + \lambda_{\text{mask}} L_{\text{mask}} + \lambda_{\text{dice}} L_{\text{dice}} + \lambda_{\text{giou}} L_{\text{giou}} + \lambda_{\text{L1}} L_{\text{L1}}.
\end{equation}

It should be noted that our model can be trained to maximize referring scores of every reasoning step if we have referent annotation for each concept node.

\section{Experiments}
\noindent\textbf{Datasets.} We evaluate EventRR and other methods on 4 RVOS benchmarks. Refer-Youtube-VOS \cite{youtube_rvos} contains 3.9k videos with 15k expressions for 7.4k objects. A2D-Sentences \cite{a2ds} contains 3.7k videos with 6.6k expressions for 4.8k objects. JHMDB-Sentences \cite{a2ds} contains 0.9k videos with 0.9k expressions for 0.9k objects, respectively. DAVIS17-RVOS \cite{davis17} contains 90 videos with 1.5k expressions for 205 objects.

\noindent\textbf{Metrics.}  Following current works \cite{referformer, html}, we use the region similarity ($\mathcal{J}$), contour accuracy ($\mathcal{F}$) and their mean value ($\mathcal{J}\&\mathcal{F}$) to evaluate Refer-Youtube-VOS and DAVIS17-RVOS. For A2D-Sentences, we report results of Precision@K (P@K), Overall IoU (oIoU), and Mean IoU (mIoU). P@K computes the percentage of predictions whose IoU is larger than K, where K is set to 0.5:0.1:0.9. Since our model outputs only one mask with the largest referring score, for fair comparison, we do not include mean Average Precision (mAP). For JHMDB-Sentences, results of oIoU and mIoU are reported.

\subsection{Implementation Details} 
\noindent\textbf{Model Details.} The visual backbones we used include ResNet\cite{resnet}, Swin Transformer \cite{swin}, and Video Swin Transformer\cite{video_swin}. The vocabulary embedding is initialized as the pre-trained vocabulary embedding in AMRBART-Large\cite{bai2022graph}. The model dimension is set to $d=256$. We set the number of frame decoder layers, SWQ layers, and temporal decoder layers to 3. The sliding window size used in SWQ is set to 6. The number of queries used in the frame decoder and temporal decoder is 20. 

\noindent\textbf{Training and Inference.} During training, a clip with 12 frames is sampled from the video. All frames are then downsampled without changing the aspect ratio such that the smaller side is resized to 360 and the longer side does not exceed 640. Random horizontal flip is used where the "left" or "right" word in the expression is changed to "right" or "left" if the video is flipped. Random crop or other color-related augmentation are not used since they may corrupt appearance or other visual context information which are referred in the expression.  We set $\lambda_{mask}=2$, $\lambda_{dice}=5$, $\lambda_{reason}=2$, $\lambda_{giou}=2$, $\lambda_{L1}=2$ in all experiments. Following the common setting of \cite{referformer, soc, html, spectrum}, we first pre-train our model by setting $T=1$ on RefCOCO\cite{refcoco,refcocog}. More details about training hyperparameters are included in the supplementary. 

During inference, all frames are downsampled to 360p. Following\cite{referformer, html}, we directly use the model trained on Ref-Youtube-VOS to infer Ref-DAVIS17, and use the model trained on A2D-Sentences to infer JHMDB-Sentence.
\subsection{Comparison with SOTA methods} 
\noindent\textbf{Ref-Youtube-VOS \& DAVIS17-RVOS.} On Ref-Youtube-VOS, EventRR is compared with other methods using four different visual backbones. As shown in Table\ref{tab:youtube_rvos}, our model achieves state-of-the-art performance in terms of the $\mathcal{J}\&\mathcal{F}$ score, which is higher than the SOTA method HTML\cite{html} by 0.3, 0.9, 1.4 points. On DAVIS17-RVOS, as shown in Table\ref{tab:youtube_rvos}, EventRR also achieves SOTA $\mathcal{J}\&\mathbf{F}$, $\mathcal{J}$, and $\mathcal{F}$ performance. In terms of $\mathcal{J}\&\mathbf{F}$, EventRR surpasses HTML\cite{html} by 0.7, 0.9 points when using ResNet-50, Swin-L as backbone, respectively. The performance consistently improves by using stronger visual backbones, which shows the generality of EventRR. 

\noindent\textbf{A2DS-Sentences \& JHMDB-Sentences.} 
Experiment results on A2D-Sentences are shown in Table\ref{tab:a2ds}. For fair comparison, we use the Video-Swin-T and Video-Swin-B as the visual backbone. In terms of P@0.6-0.9 and oIoU, EventRR achieves new state-of-the-art performance. In particular, it surpasses SOC \cite{soc} on oIoU about 0.5 points. Moreover, EventRR is superior in terms of P@0.9, which means it can generate higher-quality segmentation maps than other methods. We also evaluate EventRR with a larger spatial backbone Swin-B, which also attains superior performance. On JHMDB-Sentences \cite{a2ds}, using Video-Swin-T, EventRR achives 72.3/73.1 for mIoU and oIoU which is much better than SOTA SgMg \cite{spectrum} (71.7/72.8) and SOC \cite{soc} (71.6/72.7). Using Video-Swin-B, it achives 73.2/74.1 for mIoU and oIoU which is also much better than SOTA SgMg \cite{spectrum} (72.5/73.7) and SOC \cite{soc} (72.3/73.6).

\noindent\textbf{Vocabulary Embedding.}
Moreover, current methods HTML\cite{html}, Referformer\cite{referformer}, and SOC\cite{soc} use RoBERTa\cite{roberta} as the textual backbone, which inputs the expression into a stack of self-attention layers to encode contextualized information. EventRR simply uses the vocabulary embedding of AMRBART-Large, without the need to encode any contextualized features of the expression or REG. 

\begin{table}[htb]
\centering
\begin{tabular}{l|c| c c c | c c c}
    \toprule
    \multirow{2}{*}{Method} & \multirow{2}{*}{Backbone} & \multicolumn{3}{c}{Refer-Youtube-VOS} &\multicolumn{3}{c}{DAVIS17-RVOS}\\
     & & $\mathcal{J}\&\mathcal{F}$ & $\mathcal{J}$ & $\mathcal{F}$ & $\mathcal{J}\&\mathcal{F}$ & $\mathcal{J}$ & $\mathcal{F}$\\
    \hline
    CMSA ~\cite{ye2019cmsa} & ResNet-50        & 34.9 & 33.3 & 36.5 & 34.7 & 32.2 & 37.2 \\
    URVOS ~\cite{seo2020urvos} & ResNet-50 & 47.2 & 45.3 & 49.2 & 51.5 & 47.3 & 56.0 \\
    ReferFormer~\cite{referformer} & ResNet-50 & 55.6 & 54.8 & 56.5 & 58.5 & 55.8 & 61.3\\
    HTML~\cite{html}       & ResNet-50 & 57.8 & 56.5 & 59.0 & 59.5 & 56.6 & 62.4\\
    \rowcolor{gray!10} \textbf{EventRR}  & ResNet-50 & \textbf{59.2} & \textbf{57.6} & \textbf{60.8} & \textbf{60.2} & \textbf{57.3} & \textbf{63.1}\\ 
    \hline
    ReferFormer ~\cite{referformer} & Swin-L & 62.4 & 60.8 & 64.0 & 60.5 & 57.6 & 63.4\\
    HTML ~\cite{html} & Swin-L & 63.4 & 61.5 & 65.3 & 61.6 & 58.9 & 64.4\\
    \rowcolor{gray!10} \textbf{EventRR} & Swin-L & \textbf{64.8} & \textbf{62.8} & \textbf{66.7}& \textbf{62.7} & \textbf{59.7} & \textbf{65.7} \\
    \hline
    ReferFormer ~\cite{referformer} & Video-Swin-T & 59.4 & 58.0 & 60.9 & - & -  & -\\
    R2VOS~\cite{r2vos}& Video-Swin-T & 61.3 & 59.6 & 63.1 & - & - & - \\
    HTML ~\cite{html} & Video-Swin-T & 61.2 & 59.5 & 63.0 & - & - & -\\
    SgMg ~\cite{spectrum} & Video-Swin-T & 62.0 & 60.4 & 63.5 & 61.9 & 59.0 & 64.8\\
    TCD ~\cite{tang2023temporal} & Video-Swin-T& 62.3 & 60.5 & 64.0 & 62.2 & 59.3 & 65.0\\
    SOC ~\cite{soc}& Video-Swin-T & 62.4 & 61.1  & 63.7 & 63.5 & 60.2 & 66.7 \\
    \rowcolor{gray!10} \textbf{EventRR} & Video-Swin-T & \textbf{64.4} & \textbf{62.5} & \textbf{66.3}& \textbf{64.4} & \textbf{61.5} & \textbf{67.4} \\
    \hline
    ReferFormer ~\cite{referformer} & Video-Swin-B & 62.9 & 61.3 & 64.6& 61.1 & 58.1 & 64.1\\
    HTML ~\cite{html} & Video-Swin-B& 63.4 & 61.5 & 65.2 & 62.1 & 59.2 & 65.1\\
    SgMg ~\cite{spectrum} & Video-Swin-B & 65.7 & 63.9 & 67.4 & 63.3 & 60.6 & 66.0\\
    TCD ~\cite{tang2023temporal} & Video-Swin-B & 65.8 & 63.6 & 68.0 & 64.6 & 61.6 & 67.6\\
    SOC ~\cite{soc}&Video-Swin-B & 66.0 & 64.1  & 67.9 & 64.2 & 61.0 & 67.4 \\
    \rowcolor{gray!10} \textbf{EventRR} & Video-Swin-B & \textbf{66.9} & \textbf{65.0} & \textbf{68.8}& \textbf{65.4} & \textbf{62.7} & \textbf{68.1} \\
\bottomrule
\end{tabular}
\caption{Comparison with state-of-the-art methods on Refer-Youtube-VOS\cite{youtube_rvos} and DAVIS17-RVOS\cite{davis17}. * means training from scratch.}
\label{tab:youtube_rvos}
\end{table}

\begin{table*}[t]
\centering
\begin{tabular}{l | c | c c c c c |c c}
\toprule
\multirow{2}{*}{Method} & \multirow{2}{*}{Backbone} & \multicolumn{5}{c |}{Precision} & \multicolumn{2}{c}{IoU}\\
 & & P@0.5 & P@0.6 & P@0.7 & P@0.8 & P@0.9  &Overall & Mean\\
\midrule
Hu \textit{et al.}. ~\cite{hu16} & VGG-16  & 34.8 & 23.6 & 13.3 & 3.3 & 0.1 & 47.4 & 35.0 \\
CMSA + CFSA ~\cite{cmfsa} & ResNet-101  & 48.7 & 43.1 & 35.8 & 23.1 & 5.2 & 61.8 & 43.2 \\
CMPC-V ~\cite{cmpc} & I3D & 65.5 & 59.2 & 50.6 & 34.2 & 9.8  & 65.3 & 57.3\\
ClawCraneNet ~\cite{liang2021clawcranenet} & ResNet-50/101  & 70.4 & 67.7 & 61.7 & 48.9 & 17.1 & 63.1 & 59.9 \\
EventRR & Swin-B* & \textbf{79.1} & \textbf{76.0} & \textbf{69.2} & \textbf{54.2} & \textbf{20.1}  & \textbf{76.4} & \textbf{67.0} \\
\hline
MTTR ~\cite{mttr} & Video-Swin-T*  & 75.4 & 71.2 & 63.8 & 48.5 & 16.9  & 72.0 & 64.0\\
ReferFormer  ~\cite{referformer} & Video-Swin-T* & 76.0 & 72.2 & 65.4 & 49.8 & 17.9  & 72.3 & 64.1 \\
SOC ~\cite{soc} & Video-Swin-T* & 79.0 & 75.6 & 68.7 & 53.5 & 19.5  & 74.7 & 66.9 \\
\rowcolor{gray!10} EventRR & Video-Swin-T* & \underline{78.7} & \textbf{75.8} & \textbf{68.8} & \textbf{53.8} & \textbf{19.8}  & \textbf{75.2} & \underline{66.4} \\
\hline
ReferFormer  ~\cite{referformer} & Video-Swin-T & 82.8 & 79.2 & 72.3 & 55.3 & 19.3  & 77.6 & 69.6 \\
SOC ~\cite{soc} & Video-Swin-T & 83.1 & 80.6 & 73.9& 57.7 & 21.8  & 78.3 & 70.6 \\
HTML ~\cite{html} & Video-Swin-T & 82.2 & 79.2 & 72.3& 55.3 & 20.1  & 77.6 & 69.2 \\
\rowcolor{gray!10} EventRR & Video-Swin-T & \textbf{84.2} & \textbf{81.3} & \textbf{74.5} & \textbf{58.1} & \textbf{22.3}  & \textbf{78.8} & \textbf{71.2} \\
\hline
ReferFormer  ~\cite{referformer} & Video-Swin-B & 83.1 & 80.4 & 74.1 & 57.9 & 21.2  & 78.6 & 70.3 \\
SOC ~\cite{soc} & Video-Swin-B & 85.1 & 82.7 & 76.5& 60.7 & 25.2  & 80.7 & 72.5 \\
HTML ~\cite{html} & Video-Swin-B & 84.0 & 81.5 & 75.8 & 59.2 & 22.8  & 79.5 & 71.2 \\
\rowcolor{gray!10} EventRR & Video-Swin-B & \textbf{86.4} & \textbf{83.5} & \textbf{77.2} & \textbf{60.9} & \textbf{25.2}  & \textbf{81.4} & \textbf{73.1} \\

\bottomrule
\end{tabular}
\caption{Comparison with the state-of-the-art methods on A2D-Sentences\cite{a2ds}. * means training from scratch. Due to space limitation, we put results on JHMDB-Sentences\cite{a2ds} to Supplementary.}
\label{tab:a2ds}
\end{table*}
\subsection{Ablations}
\begin{table}[ht]
\centering
\begin{minipage}{0.45\linewidth}
    \centering
        \caption{Ablation of OCA and TRCA in TCRR.}
        \label{tab:compont_ab}
        \begin{tabular}{l c c c c}
        \toprule
        OCA & TRCA & P@0.5 & oIoU & mIoU\\
        \midrule
        & & 21.1 & 24.9 & 22.1 \\
        \checkmark &   & 77.3  &  74.3 & 65.4 \\
        &\checkmark   & 73.2  & 71.5 & 61.1 \\
        \checkmark &\checkmark   &  \textbf{79.1} & \textbf{76.4} & \textbf{67.0} \\
        \bottomrule
        \end{tabular}
\end{minipage}
\hfill
\begin{minipage}{0.45\linewidth}
    \centering
        \caption{Ablation of cross-modal information flow design in BCMF.}
        \label{tab:fusion_ab}    
        \begin{tabular}{l c c c}
        \toprule
        Fusion Strategy & P@0.5 & oIoU & mIoU \\
        \midrule
        w/o BCMF  & 52.2 & 60.1  & 43.3 \\ 
        w/o T2V   & 53.3 & 62.1  & 46.8 \\ 
        w/o V2T   & 75.2 & 73.2 & 64.2  \\
        Both &  \textbf{79.1} & \textbf{76.4} & \textbf{67.0} \\
        \bottomrule
        \end{tabular}
\end{minipage}
\end{table}

\begin{table}[ht]
\centering
\begin{minipage}{0.45\linewidth}
    \centering
        \caption{Number of decoder queries.}
        \label{tab:nq_ab}
        \begin{tabular}{l c c c}
        \toprule
         Query Number & P@0.5 & oIoU & mIoU \\
        \midrule
        10   & 78.6 & 75.6  & 66.1 \\ 
        20   &  \textbf{79.1} & \textbf{76.4} & \textbf{67.0} \\
        50   &  77.1 & 74.2 & 65.4 \\
        \bottomrule
        \end{tabular}
\end{minipage}
\hfill
\begin{minipage}{0.45\linewidth}
    \centering
        \caption{Initialization of vocabulary embedding.}
        \label{tab:vocab_ab}  
        \begin{tabular}{l c c c}
        \toprule
         Vocabulary & P@0.5 & oIoU & mIoU \\
        \midrule
        random   & 75.7 &  73.0  & 63.5 \\ 
        AMRBART   &  \textbf{79.1} & \textbf{76.4} & \textbf{67.0} \\
        \bottomrule
        \end{tabular}
\end{minipage}
\end{table}

\begin{table}[ht]
\centering
\begin{minipage}{0.45\linewidth}
    \centering
        \caption{Effect of ReferPE.}
        \label{tab:pe_ab}
        \begin{tabular}{l c c c}
        \toprule
         Position Encoding & P@0.5 & oIoU & mIoU \\
        \midrule
        w/o ReferPE & 77.2 & 74.5  & 65.1 \\ 
         w/ ReferPE & \textbf{79.1} & \textbf{76.4} & \textbf{67.0} \\
        \bottomrule
        \end{tabular}
\end{minipage}
\hfill
\begin{minipage}{0.45\linewidth}
    \centering
        \caption{Training clip size.}
        \label{tab:clip_ab}
        \begin{tabular}{l c c c}
        \toprule
        Frame Number  & P@0.5 & oIoU & mIoU \\
        \midrule
          3   &  76.4 &  72.3 & 64.5 \\
          6   &  78.1 & 74.6 & 66.0 \\
          12   & \textbf{79.1} & \textbf{76.4} & \textbf{67.0} \\
          18   & 78.8 & 76.2 & 66.8 \\
        \bottomrule
        \end{tabular}
\end{minipage}
\end{table}

We carry out extensive ablation experiments to study the effects of different modules in EventRR. All ablation experiments use Swin-B as the backbone and report P@0.5, oIoU, and mIoU on A2D-Sentences. All model variants use the same training setting as in the main experiment.

\noindent\textbf{TCRR Component Analysis.} OCA and TRCA compute referring scores based on aligning different information between two modalities. To verify our design principle and show the effectiveness of TCRR, we remove OCA or TRCA in two different variants. Moreover, we add another variant that totally removes BUTTR and randomly chooses one of the temporal queries as referent. Its performance can be seen as a lower bound of the reasoning performance. The ablation results are shown in Table \ref{tab:compont_ab}.

First, compared with the naive random sampling, adding either component can prominently improve performance. This directly demonstrates that the referring scores output by TCRR can concentrate on the referent, which verifies the effectiveness of OCA and TRCA.

Second, when using both components, performance is much higher than using either one of them. This shows that OCA and TRCA can work collaboratively to improve reasoning performance.

Third, when OCA is removed, P@0.5, oIOU, and mIOU drop by 7.45\%, 6.41\%, and 8.80\%, respectively. When TRCA is removed, P@0.5, oIoU, and mIoU drop by 2.2\%, 2.7\%, and 2.3\%, respectively. To some extent, this shows OCA is more effective than TRCA. For example, in the "$\text{walk}\stackrel{\text{:ARG0}}{\longrightarrow}\text{man}$" reasoning step, TRCA is designed to answer "who is the :ARG0 of walk?", while OCA is designed to directly answer "who is the man?". Ignoring other descriptive information and answering directly may be more effective when there are very few objects in the video. 

\begin{figure*}[thtp]
    \centering
    \begin{subfigure}[b]{\textwidth}
        \includegraphics[width=0.5\textwidth]{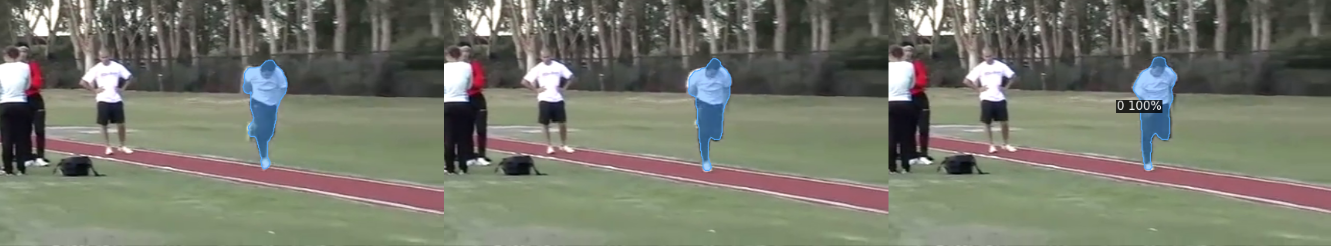}
        \includegraphics[width=0.5\textwidth]{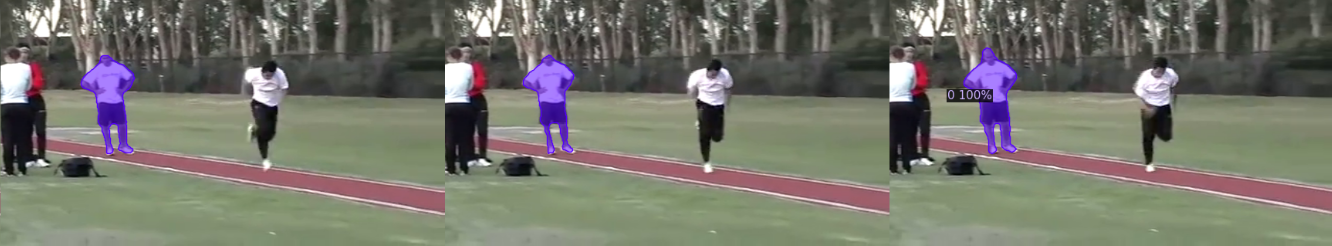}
        \caption{man in white shirt running on the track}
        \label{fig:bowling_a}
    \end{subfigure}
    \begin{subfigure}[b]{\textwidth}
        \includegraphics[width=0.5\textwidth]{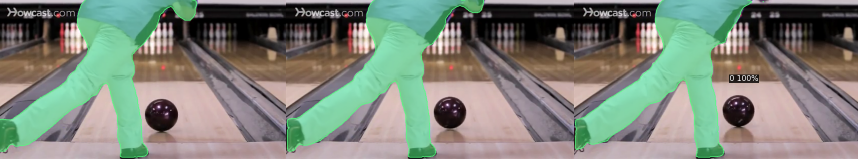}
        \includegraphics[width=0.5\textwidth]{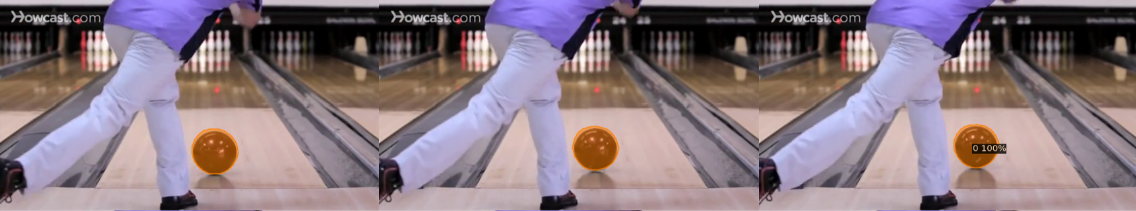}
        \caption{man throwing a bowling ball}
        \label{fig:bowling_c}
    \end{subfigure}
    \begin{subfigure}[b]{\textwidth}
        \includegraphics[width=0.5\textwidth]{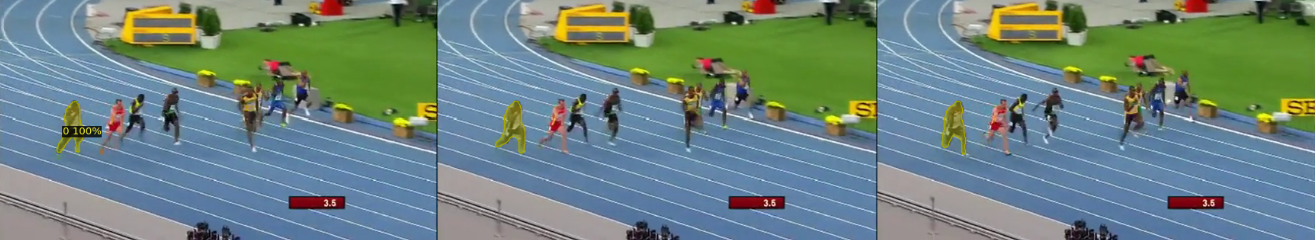}
        \includegraphics[width=0.5\textwidth]{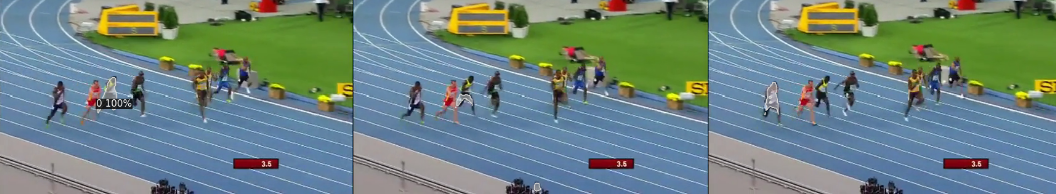}
        \caption{the first man from the bottom is running}
    \end{subfigure}
    \caption{Predictions of EventRR (left) and SOC\cite{soc} (right) for different test samples}
    \label{fig:viz}
\end{figure*}

\noindent\textbf{Information flow in BCMF.} As shown in Table \ref{tab:fusion_ab}, removing Text to Vision transfer (T2V) prominently drops the performance. This is due to that without the guidance of REG graph information, the frame-level and video-level summarization would not highlight relevant objects. The absence of Vision to Text transfer (V2T) also yields performance drop. 

\noindent\textbf{Number of decoder queries.} As shown in Table \ref{tab:nq_ab}, using more queries in frame decoder and temporal decoder increases performance. But the performance saturates after some threshold. This is due to that by using more queries, the reasoning module must account for more objects at each reasoning step. Also, utilization \cite{mask_pilot} of object queries will drop which leads to redundant information.

\noindent\textbf{Vocabulary Embedding Initialization.} As shown in Table \ref{tab:vocab_ab}, by initializing the vocabulary embedding as in pre-trained AMRBART-Large\cite{bai2022graph}, EventRR can benefit from the knowledge in the pre-training AMR corpus\cite{amr3}.

\noindent\textbf{Effect of ReferPE.}. As shown in Table \ref{tab:pe_ab}, injecting the depth-aware ReferPE to concept features can improve performance, which verifies our design principle.

\noindent\textbf{Training clip size.} As shown in Table \ref{tab:clip_ab}, by increasing the training clip size, the model can better learn the correspondance between long context temporal visual information and referring information.

\subsection{Qualitative Comparison}

To show the reasoning performance of EventRR, we compare its predictions with the current sequence-based SOC\cite{soc}. The visualizations are implemented based on Detectron2\cite{detectron2}. As shown in Figure \ref{fig:viz}, SOC fails on samples where the expression contains multiple objects. Since sequence-based methods take contextualized features as queries, they may not be able to distinguish the referent token from the context token. For example, in Figure \ref{fig:bowling_c}, "bowling" appears in the sentence, but the referent should be the "man". Moreover, in Figure \ref{fig:bowling_a}, there are multiple person who wear white shirt, while SOC identifies the standing person instead of the running person.
\section{Conclusion}

In this paper, we first delve into representing event attributes and event-event relations in video-referring expressions. The highly expressive Referential Event Graph (REG) is proposed to help RVOS models understand the semantics in a video-referring expression. Guided by REG, we propose the Event Referential Reasoning (EventRR) framework, where the Temporal Concept-Role Reasoning (TCRR) is designed to explicitly use the semantic structure in REG to identify the referent. Comprehensive experiments across four RVOS datasets showcase its state-of-the-art performance. Additionally, we conducted numerous ablation studies to validate the design principles of each EventRR component.

\clearpage
{
\small
\bibliographystyle{splncs04}
\bibliography{main}
}

\clearpage
\input{sec/X_suppl}

\end{document}

%% file: sec/X_suppl.tex
\clearpage
\setcounter{page}{1}
\appendix

\renewcommand\thefigure{\Alph{section}\arabic{figure}}  
\setcounter{figure}{0} 
\renewcommand\thetable{\Alph{section}\arabic{table}}  
\setcounter{table}{0}

\definecolor{codegreen}{rgb}{0,0.5,0}
\definecolor{codeblue}{rgb}{0.25,0.5,0.5}
\definecolor{codegray}{rgb}{0.6,0.6,0.6}


\section{Implementation Details}
We optimize the model using AdamW \cite{adamw}, with base learning rate set to $1\times 10^{-4}$, visual backbone and vocabulary embedding learning rate set to $1\times 10^{-5}$, and weight decay set to $1\times 10^{-4}$. Following Referformer\cite{referformer}, AMRefer is firstly pretrained on the the combined Referring Image Object Segmentation (RIOS) datasets of RefCOCO \cite{refcoco}, RefCOCOg\cite{refcocog}, and RefCOCO+ \cite{refcoco} by setting $T=1$. The model is pretrained for 12 epochs, where the learning rate decays by 0.1 at epoch 8 and epoch 10. Then, on Ref-Youtube-VOS and A2D-Sentences, the model is finetuned for 6 epochs, where the learning rate decays by 0.1 at epoch 3 and epoch 5. Our model is trained on 8 A800 GPUs with a batch size of 4 during pretraining and 2 during finetuning. During training from scratch, the model is trained for 20 epochs, where the batch size is set to 2, and the learning rate decays by 0.1 at epoch 16 and epoch 18.

\section{Comparison with state-of-the-arts on JHMDB-Sentences}
We also compare our EventRR with existing methods on JHMDB-Sentences\cite{a2ds} and the results are shown in Tab.\ref{tab:jhmdb}. Following ReferFormer\cite{referformer}, we directly evaluate the model trained on A2D-Sentences without any finetune. Compared with other methods, EventRR also achieves state-of-the-art performance across different backbone and training settings.
\begin{table}[htb]
 \centering
 \renewcommand{\arraystretch}{1.1}
    \resizebox{0.6\columnwidth}{!} {\begin{tabular}{l | c |c c c c}
    \toprule
    Method & Backbone & oIoU & mIoU \\
    \hline
    CMPC-V ~\cite{cmpc} & I3D & 61.6 & 61.7\\
    ClawCraneNet ~\cite{liang2021clawcranenet} & ResNet-50/101  & 64.4 & 65.6 \\
    MTTR ~\cite{mttr} & Video-Swin-T & 70.1 & 69.8\\
    ReferFormer  ~\cite{referformer} & Video-Swin-T*  & 70.0 & 69.3 \\
    SOC  ~\cite{soc} & Video-Swin-T*  & 70.7 & 70.1 \\
    \rowcolor{gray!10} EventRR  &  Video-Swin-T*  & \textbf{71.1} & \textbf{70.2}\\
    \hline
    ReferFormer ~\cite{referformer} & Video-Swin-T & 71.9 & 71.0\\
    SgMG ~\cite{spectrum} & Video-Swin-T & 72.8 & 71.7 \\
    SOC ~\cite{soc} & Video-Swin-T & 72.7 & 71.6\\
    \rowcolor{gray!10} EventRR  &  Video-Swin-T  & \textbf{73.0} & \textbf{71.9}\\
    \hline
    ReferFormer ~\cite{referformer} & Video-Swin-B & 73.0 & 71.8\\
    SgMG ~\cite{spectrum} & Video-Swin-B & 73.7 & 72.5  \\
    SOC ~\cite{soc} & Video-Swin-B & 73.6 & 72.3\\
    \rowcolor{gray!10} EventRR  &  Video-Swin-B  & \textbf{73.9} & \textbf{72.6}\\    
    \bottomrule
    \end{tabular}}
    \caption{Comparison with state-of-the-art methods on JHMDB-Sentences\cite{a2ds}. * means training from scratch.}
    \label{tab:jhmdb}
\end{table}
\vspace{-8mm}

\section{Referential Event Graph}\label{sec:semantic_extraction}
To represent the compositional semantic structure in a video-referring expression, we first use a transition-based AMR parser \cite{amr_parser} to get the AMR parsing graph of the expression. 
However, since the AMR annotation process \cite{amr} does not require annotators to specify the focus of a sentence, the root node of the generated AMR graph does not often correspond to the referent main concept, which does not induce a clear reasoning process. Moreover, since co-reference may appear, rings will appear in the adjacency-based graph representation, which fails topological sort. Moreover, files of REG parsing result on A2D-Sentences\cite{a2ds} and Refer-Youtube-VOS\cite{youtube_rvos} are included in supplementary materials. Detailed implementations can be found in the code. 

To solve the first issue, we aim to use the syntactic structure of the sentence to choose the referent concept. In particular, we first use the Stanford CoreNLP PipeLine \cite{manning2014stanford} to get the universal dependency and position-of-speech (POS) parsing results for the sentence. Then the referent token is chosen as the first token whose POS tag is NN/NNS/NNP/NNPS or the last token in a 'nn:compound' dependency relation. After locating the referent token, we use the concept-token alignment provided by the transition-based AMR parser to find the set of concepts that are parsed from the refernt token. If there are more than one concepts parsed from the token, we choose the concept whose lemma has the largest common string with the referent token. We found this strategy can succeed for 98.3\%, 99.9\% of all sentences in MeViS \cite{mevis} and Refer-Youtube-VOS \cite{youtube_rvos}, respectively. If no referent token can be found or no concept is parsed from the referent token, we just choose the concept with the smallest token alignment index as the referent concept. Finally, we follow the AMR specification\cite{amr, amr_website} to change the root of the graph to the target referent concept.
To solve the second issue, we use depth-first search to incrementally build the directed acyclic graph starting from the root. For each child concept, if the ascendants of the parent concept contains the child concept, co-reference appears and we revert the edge direction and change the role to its inverse role\cite{amr_website, amr}. 

\begin{figure}[t]
\begin{center}
\includegraphics[width=\linewidth]{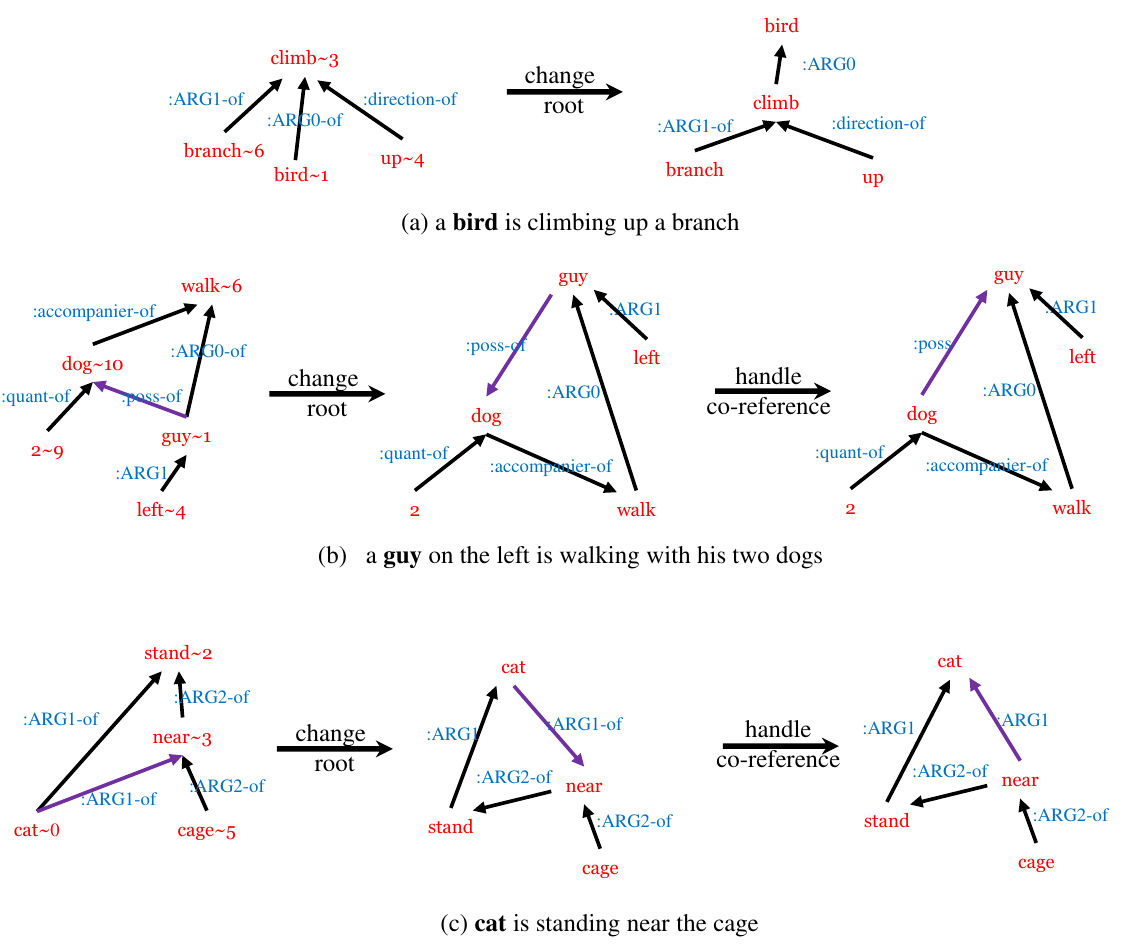}
\end{center}
\vspace{-6mm}
\caption{Examples of transform from AMR to REG. The co-reference arrow is highlighted as purple. The referent token is highlighted in bold format. $\sim$ in the first graph denotes the concept-token alignment provided by the AMR parser. The token index starts from 0. (a) a bird is climbing up a branch. (b) a guy on the left is walking with his two dogs. (c) cat is standing near the cage.}
\label{fig:parser_examples}
\vspace{-4mm}
\end{figure}
For example, in Figure \ref{fig:parser_examples} (c), following the reverse edge direction, the DFS order is "cat $\Rightarrow$ stand $\Rightarrow$ near". When traversal arrives at "near", current graph is "$\text{near}\stackrel{\text{:ARG2-of}}{\longrightarrow}\text{stand}\stackrel{\text{:ARG1}}{\longrightarrow}\text{cat}$", its ascendants are "stand" and "cat". For the "$\text{cage}\stackrel{\text{:ARG2-of}}{\longrightarrow}\text{near}$" edge, since "cage" does not appear in current graph, this edge is directly added to current graph. For the "$\text{cat}\stackrel{\text{:ARG1-of}}{\longrightarrow}\text{near}$" edge, "cat" is not only a child but also an ascendant. In this case, we invert the edge direction and change the role to its inverse role \cite{amr_website}, i.e., "$\text{near}\stackrel{\text{:ARG1}}{\longrightarrow}\text{cat}$". The final graph is acyclic, and rooted by the referent concept, which induces a clear process for referent reasoning.
